\documentclass[journal]{IEEEtran}
\usepackage{setspace}
\usepackage[section]{placeins}
\usepackage{graphicx,subfigure}
\usepackage{amsmath}
\usepackage{amsfonts}
\usepackage{caption2}
\usepackage{amssymb}
\usepackage{varwidth}
\usepackage{multirow}
\usepackage{booktabs}
\usepackage[usenames, dvipsnames]{color}
\usepackage[linesnumbered,ruled,vlined]{algorithm2e}
\usepackage{array}
\usepackage{rotating}
\usepackage{epsfig,amsfonts,multirow}
\usepackage{bm}
\usepackage{cite}

\begin{document}
\title{Classification of Hyperspectral and LiDAR Data Using Coupled CNNs}
\author{Renlong~Hang,~\IEEEmembership{Member,~IEEE}, Zhu Li,~\IEEEmembership{Senior~Member,~IEEE}, Pedram Ghamisi,~\IEEEmembership{Senior~Member,~IEEE}, \\
Danfeng Hong,~\IEEEmembership{Member,~IEEE}, Guiyu Xia, and Qingshan~Liu,~\IEEEmembership{Senior~Member,~IEEE}

\thanks{This work was supported in part by the Natural Science Foundation of United States under grant 1747751, in part by the Natural Science Foundation of China under Grants 61825601, 61532009, 61906096, and 61802198, in part by the Natural Science Foundation of Jiangsu Province, China, under Grants BK20180786, BK20180788, and 18KJB520032. (Corresponding author: Zhu Li.)

R. Hang is with the Jiangsu Key Laboratory of Big Data Analysis Technology, the School of Automation, Nanjing University of Information Science and Technology, Nanjing 210044, China, and also with the Department of Computer Science and Electrical Engineering, University of Missouri-Kansas City, Missouri 64110, USA (renlong\_hang$@$163.com).

Z. Li is with the Department of Computer Science and Electrical Engineering, University of Missouri-Kansas City, Missouri 64110, USA (lizhu@umkc.edu).

P. Ghamisi is with the Helmholtz-Zentrum Dresden-Rossendorf (HZDR), Helmholtz Institute Freiberg for Resource Technology (HIF), Exploration, D-09599 Freiberg, Germany (e-mail: p.ghamisi@gmail.com).

D. Hong is with the Remote Sensing Technology Institute (IMF), German Aerospace Center (DLR), 82234 Wessling, Germany, and Signal Processing in Earth Observation (SiPEO), Technical University of Munich (TUM), 80333 Munich, Germany (e-mail: danfeng.hong@dlr.de).

G. Xia and Q. Liu are with the Jiangsu Key Laboratory of Big Data Analysis Technology, the School of Automation, Nanjing University of Information Science and Technology, Nanjing 210044, China (xiaguiyu1989@sina.com, qsliu$@$nuist.edu.cn).
}}

\maketitle

\begin{abstract}
\textcolor{blue}{This paper has been accepted by IEEE Transactions on Geoscience and Remote Sensing.} In this paper, we propose an efficient and effective framework to fuse hyperspectral and Light Detection And Ranging (LiDAR) data using two coupled convolutional neural networks (CNNs). One CNN is designed to learn spectral-spatial features from hyperspectral data, and the other one is used to capture the elevation information from LiDAR data. Both of them consist of three convolutional layers, and the last two convolutional layers are coupled together via a parameter sharing strategy. In the fusion phase, feature-level and decision-level fusion methods are simultaneously used to integrate these heterogeneous features sufficiently. For the feature-level fusion, three different fusion strategies are evaluated, including the concatenation strategy, the maximization strategy, and the summation strategy. For the decision-level fusion, a weighted summation strategy is adopted, where the weights are determined by the classification accuracy of each output. The proposed model is evaluated on an urban data set acquired over Houston, USA, and a rural one captured over Trento, Italy. On the Houston data, our model can achieve a new record overall accuracy of 96.03\%. On the Trento data, it achieves an overall accuracy of 99.12\%. These results sufficiently certify the effectiveness of our proposed model.

\end{abstract}
\begin{IEEEkeywords}
Convolutional neural networks (CNNs), hyperspectral data, Light Detection And Ranging (LiDAR) data, parameter sharing, feature fusion, decision fusion.

\end{IEEEkeywords}
\IEEEpeerreviewmaketitle

\section{Introduction}
Accurate land-use and land-cover classification plays an important role in many applications such as urban planning and change detection. In the past few years, hyperspectral data have been widely explored for this task \cite{hang2015matrix, ghamisi2017, hong2018augmented}. Compared to multispectral data, hyperspectral data have more rich spectral information, ranging from the visible spectrum to the infrared spectrum \cite{hong2019cospace}. Such information, combined with some spatial information in hyperspectral data, can generally acquire satisfying classification results \cite{ghamisi2018,he2018}. However, for urban and rural areas, there often exist many complex objects that are difficult to discriminate, because they have similar spectral responses. Thanks to the development of remote sensing technologies, nowadays, it is possible to measure different aspects of the same object on the Earth's surface \cite{ghamisi2018multisource}. Different from hyperspectral data, Light Detection And Ranging (LiDAR) data can record the elevation information of objects, thus providing complementary information for hyperspectral data. For instance, if the building roof and the road are both
made up of concrete, it is very difficult to distinguish them using only hyperspectral data since their spectral responses are similar. However, LiDAR data can accurately classify those two classes as they have different heights. On the contrary, LiDAR data cannot differentiate between two different roads, which are made up of different materials (e.g., asphalt and concrete), having the same height. Therefore, fusing hyperspectral and LiDAR data is a promising scheme whose performance has already been validated in the literature for land-cover and land-use classification \cite{ghamisi2018multisource,debes2014}.

In order to take advantage of the complementary information between hyperspectral and LiDAR data, a lot of works have been proposed. One widely used class of methods is based on the feature-level fusion. In \cite{pedergnana2012}, morphological extended attribute profiles (EAPs) were applied to hyperspectral and LiDAR data respectively. These profiles and the original spectral information of hyperspectral data were stacked together for classification. However, the direct stacking of these high-dimensional features inevitably results in the well-known Hughes phenomenon, especially when only a relatively small number
of training samples is available. To address this issue, principal component analysis (PCA) was employed to reduce the dimensionality. Similar to this work, many subspace-related models can be designed to fuse the extracted spectral, spatial, and elevation features \cite{zhang2016multisource,liao2015,rasti2017fusion,rasti2017,rasti2019}. For example, a graph embedding framework was proposed in \cite{liao2015}; a low-rank component analysis model was proposed in \cite{rasti2017fusion}. Different from them, Gu \textit{et al.} attempted to use multiple-kernel learning \cite{niazmardi2018} to combine heterogeneous features \cite{gu2015}. They constructed a kernel for each feature, and then combined these kernels together in a weighted summation manner. Different weights can represent the importance of different features for classification.

Besides the feature-level fusion, the decision-level fusion is another popularly adopted method. In \cite{liao2014}, spectral features, spatial features, elevation features, and their fused features were fed into the support vector machine (SVM) individually to generate four classifiers, and the final classification result was determined by them. In \cite{zhang2015ensemble}, two different fusion strategies named hard decision fusion and soft decision fusion were used to integrate the classification results from different data source. Their fusion weights were uniformly distributed. In \cite{zhong2017}, three different classifiers, including the maximum likelihood classifier, SVM, and the multinomial logistic regression, were used to classify the extracted features. The fusion weights for these classifiers were adaptively optimized by a differential evolution algorithm. Recently, a novel ensemble classifier using random forest was proposed, in which a majority voting method was used to produce the final classification result \cite{xia2018}. In summary, the difference between feature-level fusion and decision-level fusion methods lies in the phase where the fusion process happens, but both of them require powerful representations of hyperspectral and LiDAR data. To achieve this goal, one needs to spend a lot of time designing appropriate feature extraction and feature selection methods. These handcrafted features often require domain expertise and prior knowledge.

In recent years, deep learning has attracted more and more attention in the field of remote sensing \cite{zhang2016, Hang2019Cascaded}. In contrast to the handcrafted features, deep learning can learn high-level semantic features from data itself in an end-to-end manner \cite{zhu2017}. Among various deep learning models, convolutional neural networks (CNNs) gain the most attention and have been explored in various tasks. For example, in \cite{cheng2016}, CNN was applied to object detection in remote sensing images. In \cite{chen2016deep}, three CNN frameworks were proposed for hyperspectral image classification. In \cite{liu2018}, Liu \textit{et al.} used CNNs to learn multi-scale deep features for remote sensing image scene classification. Due to its powerful feature learning ability, some researchers attempted to use CNN for hyperspectral and LiDAR data fusion recently. An early attempt appears in \cite{morchhale2016}. It directly considered LiDAR data as another spectral band of hyperspectral data, and then fed the concatenated data into CNN to learn features and perform classification. In \cite{ghamisi20172}, Ghamisi \textit{et al.} tried to combine the traditional feature extraction method and CNN together. They fed the fused features to CNN for learning a higher-level representation and getting a classification result. Similarly, Li \textit{et al.} constructed three CNNs to learn spectral, spatial, and elevation features, respectively, and then used a composite kernel method to fuse them \cite{li2018}. Different from them, an end-to-end CNN fusion model was designed in \cite{chen2017deep}, which embedded feature extraction, feature fusion, and classification into one framework. Specifically, the hyperspectral and LiDAR data were directly fed into their corresponding CNNs to extract features, and then these features were concatenated together, followed by a fully-connected layer to further fuse them. Based on this two-branch framework, Xu \textit{et al.} also proposed a spectral-spatial CNN for hyperspectral data analysis and another spatial CNN for LiDAR data analysis \cite{xu2018}.

It is well-known that the performance of CNN-based models heavily depends on the number of available samples. However, in the field of hyperspectral and LiDAR data fusion, there often exists a small number of training samples. To address this issue, an unsupervised CNN model was proposed in \cite{zhang2018} based on the famous encoder-decoder architecture \cite{long2015}. Specifically, it first mapped the hyperspectral data into a hidden space via an encoding path, and then reconstructed the LiDAR data with a decoding path. After that, the hidden representation in the encoding path can be considered as fused features of hyperspectral and LiDAR data. Nevertheless, there still exist some issues. For examples, the loss of supervised information from labeled samples will lead to a suboptimal feature representation; it also needs to design another network to classify the learned representation, which will increase the computation complexity. In this paper, we propose a supervised model to fuse hyperspectral and LiDAR data by designing an efficient and effective CNN framework. Similar to \cite{chen2017deep}, we also use two CNNs but with a more efficient representation. We use three convolutional layers with small kernels (i.e., $3\times3$), and two of them share parameters. Besides the output layer, we do not use any fully-connected layers. The major contributions of this paper are summarized as follows.

\begin{enumerate}
  \item In order to sufficiently fuse hyperspectral and LiDAR data, two coupled CNNs are designed. Compared to the existing CNN-based fusion models, our model is more efficient and effective. The coupled convolution layers can reduce the number of parameters, and more importantly, guide the two CNNs learn from each other, thus facilitating the following feature fusion process.
  \item In the fusion phase, we simultaneously use feature-level and decision-level fusion strategies. For the feature-level fusion, we propose summation and maximization fusion methods in addition to the widely adopted concatenation method. To enhance the discriminative ability of learned features, we add two output layers to the CNNs, respectively. These three output results are finally combined together via a weighted summation method, whose weights are determined by the classification accuracy of each output on the training data.
  \item We test the effectiveness of the proposed model on two data sets using standard training and test sets. On the Houston data, we can achieve an overall accuracy of 96.03\%, which is the best result ever reported in the literature. On the Trento data, we can also obtain very high performance (i.e., an overall accuracy of 99.12\%).
\end{enumerate}

The rest of this paper is organized as follows. Section II describes the details of the proposed model, including the coupled CNN framework, the data fusion model, and the network training and testing methods. The descriptions of data sets and experimental results are given in Section III. Finally, Section IV concludes this paper.

\section{Methodology}
\subsection{Framework of the Proposed Model}
\begin{figure*}
  \centering
  \includegraphics[scale = 0.65]{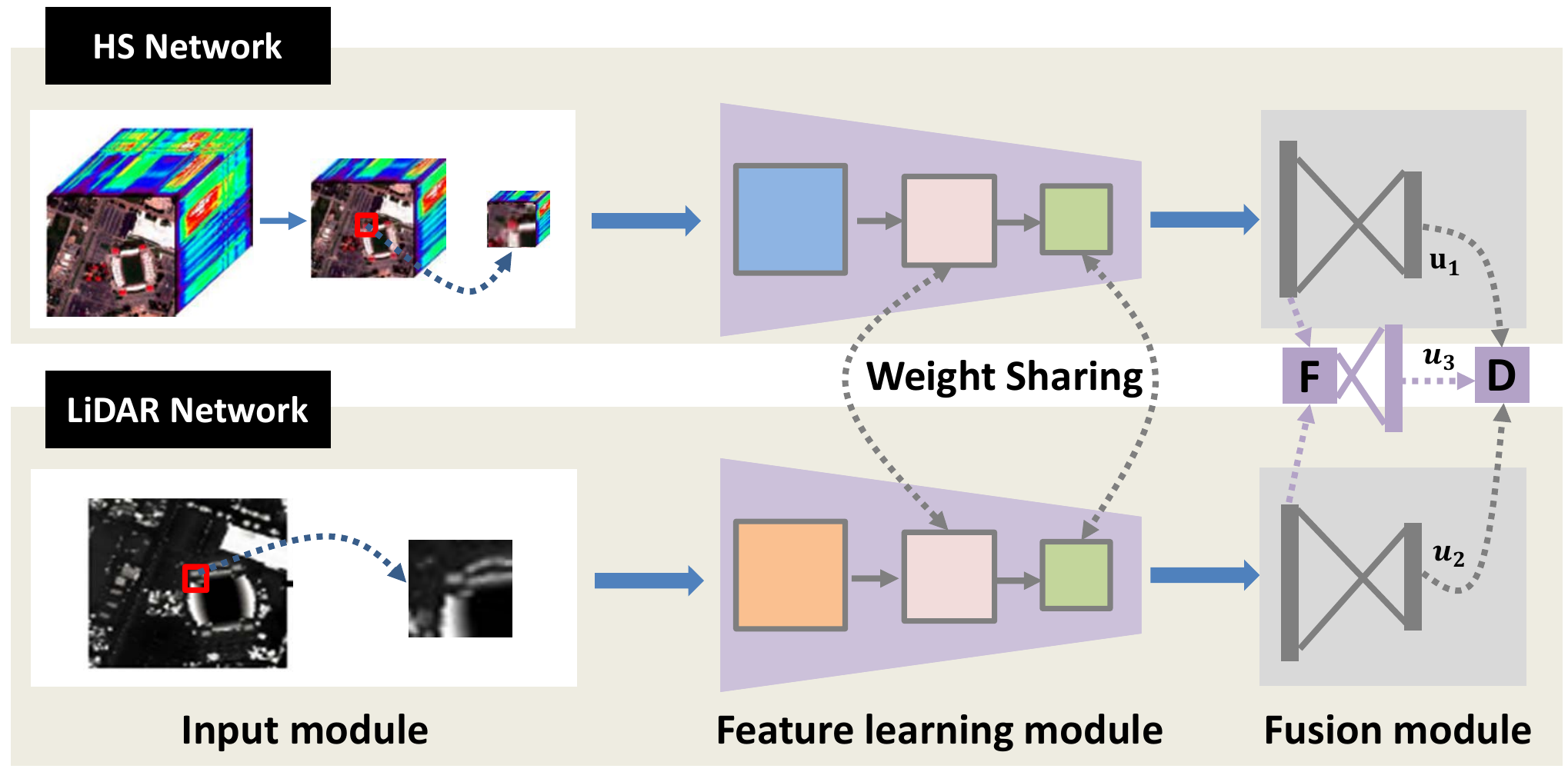}\\
  \caption{Flowchart of the proposed model.}\label{Framework}
\end{figure*}

As shown in Fig.$~$\ref{Framework}, our proposed model mainly consists of two networks: an HS network for spectral-spatial feature learning and a LiDAR network for elevation feature learning. Each of them includes an input module, a feature learning module and a fusion module. For the HS network, PCA is firstly used to reduce the redundant information of the original hyperspectral data, and then a small cube is extracted surrounding the given pixel. For the LiDAR network, we can directly extract an image patch at the same spatial position as the hyperspectral data. In the feature learning module, we use three convolutional layers, and the last two of them share parameters. In the fusion module, we construct three classifiers. Each CNN has an output layer, and their fused features are also fed into an output layer.

\subsection{Feature Learning via Coupled CNNs}
\begin{figure}
  \centering
  \includegraphics[scale = 0.45]{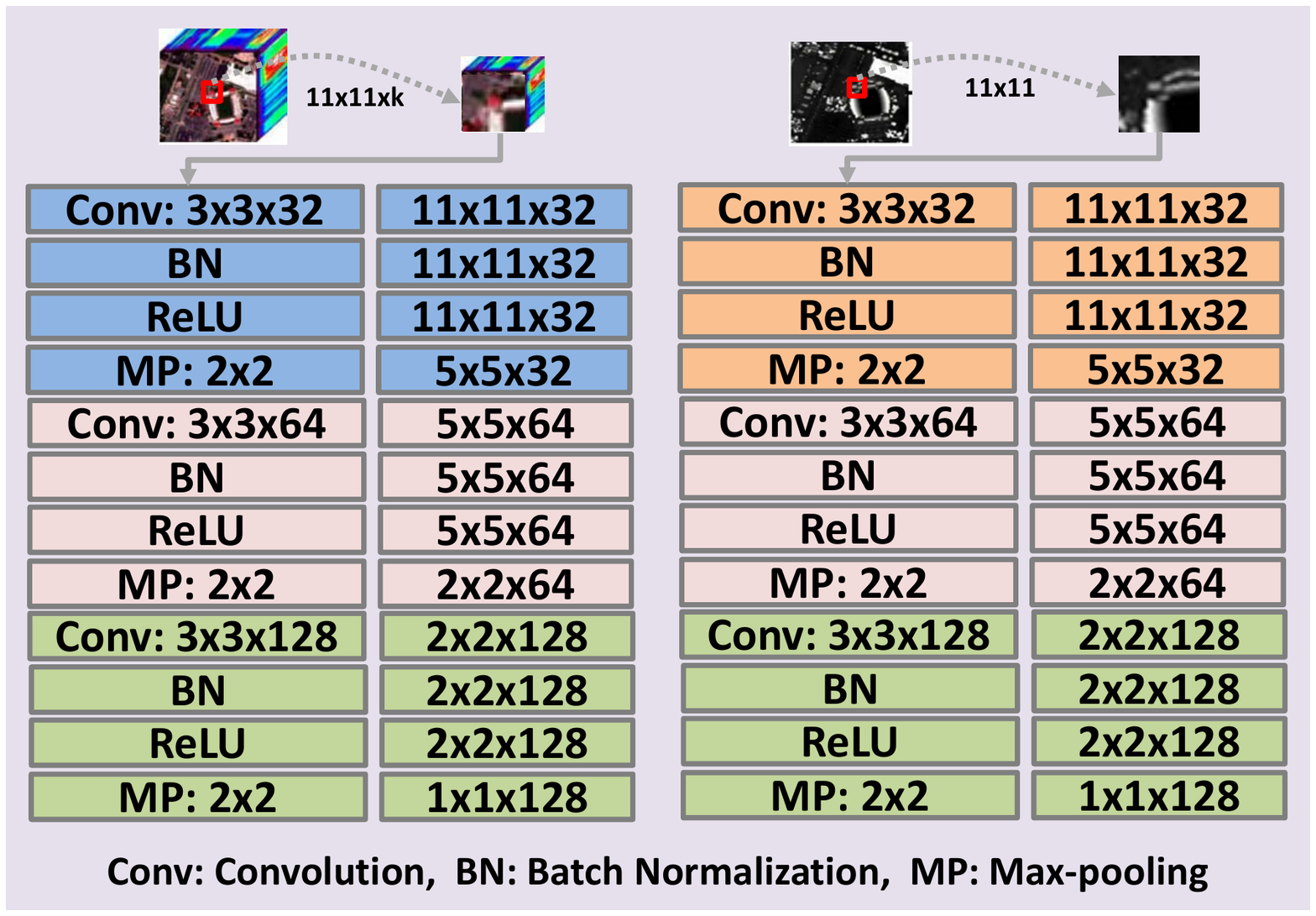}\\
  \caption{Architecture of the coupled CNNs.}\label{Structure}
\end{figure}

Given a hyperspectral image $\mathbf{X}_{h}\in \mathfrak{R}^{m\times n\times b}$ and a corresponding LiDAR image $\mathbf{X}_{l}\in \mathfrak{R}^{m\times n}$ covering the same area on the Earth's surface. Here, $m$ and $n$ represent the height and width, respectively, of the two images, and $b$ refers to the number of spectral bands of the hyperspectral image. Our goal is to sufficiently fuse the information from $\mathbf{X}_{h}$ and $\mathbf{X}_{l}$ to improve the classification performance. The same as other classification tasks, feature representation is a critical step here. Due to the effects of multi-path scattering and the heterogeneity of sub-pixel constituents, $\mathbf{X}_{h}$ often exhibits nonlinear relationships between the captured spectral information and the corresponding material. This nonlinear characteristic will be magnified when dealing with $\mathbf{X}_{l}$ \cite{ghamisi2018multisource}. It has been proved that CNNs are capable of extracting high-level features, which are usually invariant to the nonlinearities of hyperspectral \cite{yu2017,xu2018SS,zhong2018} and LiDAR data \cite{chen2017deep,he2018lidar}. Inspired from them, we design a coupled CNN framework to learn features from $\mathbf{X}_{h}$ and $\mathbf{X}_{l}$ efficiently.

The detailed architecture of the coupled CNNs is demonstrated in Fig.$~$\ref{Structure}. First of all, PCA is used to extract the first $k$ principle components of $\mathbf{X}_{h}$ to reduce the redundant spectral information. Then, for each pixel, a small cube $\textbf{\textit{x}}_{h}\in\mathfrak{R}^{p\times p\times k}$ and a small patch $\textbf{\textit{x}}_{l}\in\mathfrak{R}^{p\times p}$ centered at it are chosen from $\mathbf{X}_{h}$ and $\mathbf{X}_{l}$, respectively. According to \cite{chen2017deep} and \cite{zhang2018}, the neighboring size $p$ can be empirically set to 11. After that, $\textbf{\textit{x}}_{h}$ and $\textbf{\textit{x}}_{l}$ are fed into three convolutional layers to learn features. For the first convolutional layer, we adopt two different convolution operators (the blue box and the orange box) to obtain an initial representation of $\textbf{\textit{x}}_{h}$ and $\textbf{\textit{x}}_{l}$, respectively. This convolutional layer is sequentially followed by a batch normalization (BN) layer to regularize and accelerate the training process, a rectified linear unit (ReLU) to learn a nonlinear representation, and a max-pooling layer to reduce the data variance and the computation complexity.

For the second convolutional layer, we let the HS network and the LiDAR network share parameters. Such a coupling strategy has at least two benefits. First, it can significantly reduce the number of parameters by twice, which is very useful with small numbers of training samples. Second, it can make these two networks learn from each other. Without weight sharing, the training parameters in each network will be optimized independently using their own loss functions. After adopting the coupling strategy, the back-propagated gradients to this layer will be determined by the loss functions of both networks, which means that the information in one network will directly affect the other one. For the third convolutional layer, we also use the coupling strategy, which can further improve the discriminative ability of the learned representation from the second convolutional layer. Again, these two convolutional layers are followed by BN, ReLU, and max-pooling operators. The sizes (i.e., $3\times3$) and the number of kernels (i.e., 32, 64 and 128 sequentially) of each convolutional layer are shown at the left side under each data. Similarly, the output size (e.g., $11\times11\times32$) of each operator is shown at the right side.
%For the ($l$+1)-th convolutional layer, its output can be described as:
%\begin{equation}
%  \mathbf{x}_{j}^{l+1} = \sum_{i=1}^{N}\mathbf{x}_{i}^{l}\ast\mathbf{w}_{j}^{l+1}+b_{j}^{l+1}
%\end{equation}
%where $\mathbf{x}_{i}^{l}$ and $N$ are the $i$-th feature map and the total number of feature maps at the $l$-th layer, respectively, $b_{j}^{l+1}$ is the bias of the $j$-th filter, $\mathbf{x}_{j}^{l+1}$ and $\mathbf{w}_{j}^{l+1}$ are the $j$-th feature map and the the $j$-th filter, at the current ($l$+1)-th layer, respectively.
It is worth noting that all the convolutional layers have padding operators to make the output size the same as the input size.

\subsection{Hyperspectral and LiDAR Data Fusion }
\begin{figure}
  \centering
  \includegraphics[scale = 0.6]{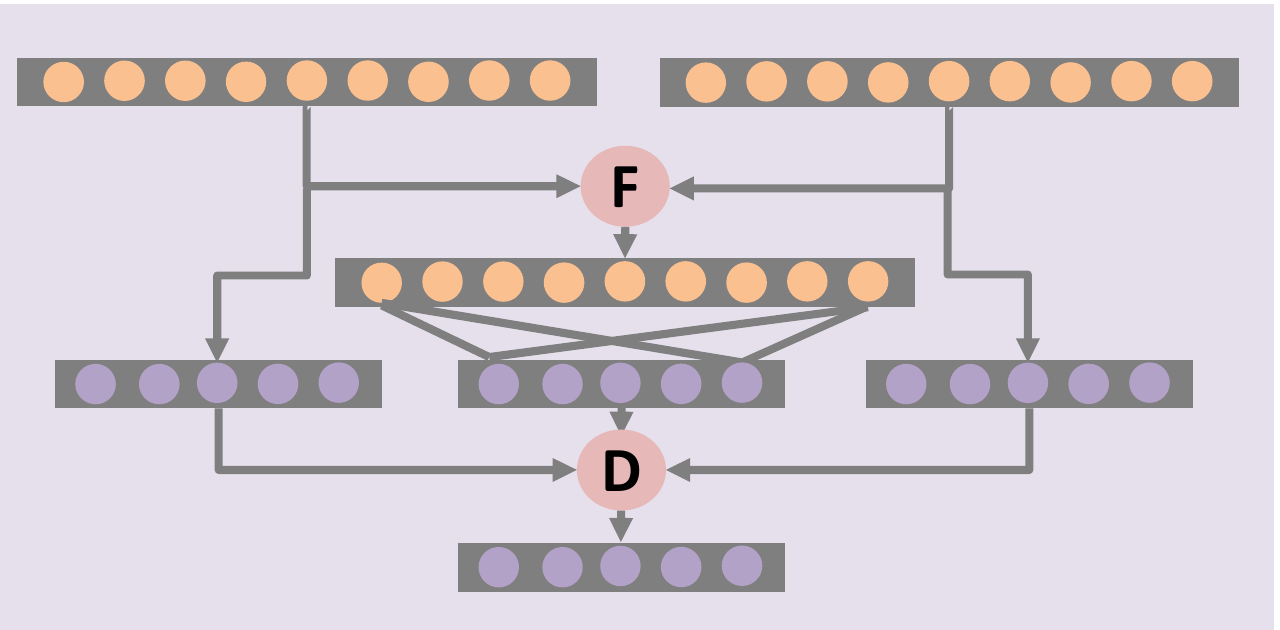}\\
  \caption{Structure of the fusion module.}\label{Fusion}
\end{figure}
After getting the feature representations of $\textbf{\textit{x}}_{h}$ and $\textbf{\textit{x}}_{l}$, how to combine them becomes another important issue. Most of the existing deep learning models \cite{chen2017deep,zhang2018,xu2018} choose to stack them together and use a few fully-connected layers to fuse them. However, fully-connected layers often contain large numbers of parameters, which will increase the training difficulty when there only exists a small number of training samples. To this end, we propose a novel combination strategy based on feature-level and decision-level fusions. Assume $\mathbf{R}_{h}\in\mathfrak{R}^{128\times1}$ and $\mathbf{R}_{l}\in\mathfrak{R}^{128\times1}$ denote the learned features for $\textbf{\textit{x}}_{h}$ and $\textbf{\textit{x}}_{l}$, respectively. As shown in Fig.$~$\ref{Fusion}, we first combine $\mathbf{R}_{h}$ and $\mathbf{R}_{l}$ to generate a new feature representation. Then, we input these three features into output layers separately. Finally, all the output layers are integrated together to produce a final result. The whole fusion process can be formulated as:
\begin{equation}
 \mathbf{O} = D[f_{1}(\mathbf{R}_{h};\mathbf{W}_{1}), f_{2}(\mathbf{R}_{l};\mathbf{W}_{2}), f_{3}(F(\mathbf{R}_{h},\mathbf{R}_{l});\mathbf{W}_{3}); \mathbf{U}]
\end{equation}
In the above equation, $\mathbf{O}\in\mathfrak{R}^{C\times1}$, where $C$ is the number of classes to discriminate, reprensents the final output of the fusion module; $D$ and $F$ are decision-level and feature-level fusions, respectively; $f_{1}$, $f_{2}$, and $f_{3}$ are three output layers connected to $\mathbf{R}_{h}$, $\mathbf{R}_{l}$, and $F(\mathbf{R}_{h},\mathbf{R}_{l})$, respectively; $\mathbf{W}_{1}\in\mathfrak{R}^{C\times128}$, $\mathbf{W}_{2}\in\mathfrak{R}^{C\times128}$,
$\mathbf{W}_{3}\in\mathfrak{R}^{C\times128}$, denote the connection weights for $f_{1}$, $f_{2}$, and $f_{3}$, respectively; $\mathbf{U}\in\mathfrak{R}^{C\times3}$ corresponds to the fusion weight for $D$.

For the feature-level fusion $F$, we use summation and maximization methods in addition to the widely used concatenation method. The summation fusion aims to compute the sum of the two representations:
\begin{equation}
  F(\mathbf{R}_{h},\mathbf{R}_{l}) = \mathbf{R}_{h} + \mathbf{R}_{l}
\end{equation}
Similarly, the maximization fusion aims at performing an element-wise maximization:
\begin{equation}
  F(\mathbf{R}_{h},\mathbf{R}_{l}) = max(\mathbf{R}_{h},\mathbf{R}_{l})
\end{equation}
Obviously, the performance of $F$ depends on its inputs $\mathbf{R}_{h}$ and $\mathbf{R}_{l}$. Therefore, we add two output layers $f_{1}$, and $f_{2}$ to supervise their learning processes. In the output phase, they can also help make decisions. The output value of $f_{1}$ can be derived as follows:
\begin{equation}\label{soft}
  \hat{\mathbf{y}}_{1} = f_{1}(\mathbf{R}_{h};\mathbf{W}_{1}) = softmax(\mathbf{W}_{1}\mathbf{R}_{h})
\end{equation}
where $softmax$ represents the softmax function. Similar to Equation$~$(\ref{soft}), we can also derive the output values $\hat{\mathbf{y}}_{2}$ and $\hat{\mathbf{y}}_{3}$ for $f_{2}$ and $f_{3}$, respectively. For the decision-level fusion $D$, we adopt a weighted summation method:
\begin{equation}\label{output}
  \mathbf{O} = D(\hat{\mathbf{y}}_{1},\hat{\mathbf{y}}_{2},\hat{\mathbf{y}}_{3};\mathbf{U})
  = \mathbf{u}_{1}\odot\hat{\mathbf{y}}_{1}+\mathbf{u}_{2}\odot\hat{\mathbf{y}}_{2}
  +\mathbf{u}_{3}\odot\hat{\mathbf{y}}_{3}
\end{equation}
where $\odot$ is an element-wise product operator, $\mathbf{u}_{1}$, $\mathbf{u}_{2}$ and $\mathbf{u}_{3}$ are three column vectors of $\mathbf{U}$, and the $i$-th element of $\mathbf{u}_{j}, j\in\{1,2,3\}$ depends on the $i$-th class accuracy acquired by the $j$-th output layer on the training data.

\subsection{Network Training and Testing}
The whole network in Fig.$~$\ref{Framework} is trained in an end-to-end manner using a given training set $\{(\textbf{\textit{x}}_{h}^{(i)},\textbf{\textit{x}}_{l}^{(i)},\mathbf{y}^{(i)})|i=1,2,\cdots,N\}$, where $N$ represents the number of training samples, and $\mathbf{y}^{(i)}$ is the ground-truth for the $i$-th sample. After a feed-forward process, we are able to obtain three outputs for each sample. Their loss values can be computed by a cross-entropy loss function. For instance, the loss value between the first output $\hat{\mathbf{y}}_{1}$ and the ground-truth $\mathbf{y}$ can be formulated as
\begin{equation}
  L_{1} = -\frac{1}{N}\sum_{i=1}^{N}[\mathbf{y}^{(i)}\mathrm{log}(\hat{\mathbf{y}}_{1}^{(i)})+
  (1-\mathbf{y}^{(i)})\mathrm{log}(1-\hat{\mathbf{y}}_{1}^{(i)})]
\end{equation}
Similarly, we can also derive $L_{2}$ and $L_{3}$ for the other two outputs.
%The final loss value $L$ is the combination of $L_{1}$, $L_{2}$ and $L_{3}$. Since our purpose is to get a powerful representation that can capture the discriminative information from hyperspectral and LiDAR data simultaneously, $L_{3}$ should have a larger combination weight than $L_{1}$ and $L_{2}$. Thus, we can empirically set $L$ as
$L_{3}$ is designed to supervise the learning process of the fused feature between hyperspectral and LiDAR data, while $L_{1}$ and $L_{2}$ are responsible for the hyperspectral and LiDAR features, respectively.
The final loss value $L$ is represented as the combination of $L_{1}$, $L_{2}$, and $L_{3}$:
\begin{equation}\label{Loss}
  L = \lambda_{1}L_{1} + \lambda_{2}L_{2} + L_{3}
\end{equation}
where $\lambda_{1}$ and $\lambda_{2}$ represent the weight parameters for $L_{1}$ and $L_{2}$, respectively. In the experiments, we empirically set them to 0.01 because it can achieve satisfactory performance. The effects of them on the classification performance will be analysed in section III-D.

%Since the proposed model aims at capturing the complementary information from hyperspectral and LiDAR data to improve the classification performance, it is necessary to set a larger weight to $L_{3}$ than the other ones. In addition, the fused feature is dependent on the extracted hyperspectral and LiDAR features. To ensure the discriminative ability of them, the weights for $L_{1}$ and $L_{2}$ can not be set too small. Based on these analyses, we empirically set $L$ as
%\begin{equation}
%  L = 0.01\times L_{1} + 0.01\times L_{2} + L_{3}
%\end{equation}
The same as most CNN models, $L$ can be optimized using a backpropagation algorithm. Note that $L_{1}$ and $L_{2}$ can also be considered as regularization terms for $L_{3}$, thus reducing the overfitting risk during the network training process.

Once the network is trained, we can use it to predict the label of each test sample. Firstly, $\mathbf{u}_{j}, j\in\{1,2,3\}$ is computed on the training set. Its $i$-th element $\mathbf{u}_{ji}$ can be derived as
%\begin{equation}
%  \mathbf{u}_{ji} = \frac{\sum_{\ell=1}^{N}\sum_{\mathbf{y}^{(\ell)}=i}\mathbf{I}(\hat{\mathbf{y}}_{j}^{(\ell)}=\mathbf{y}^{(\ell)})}
%  {\sum_{\ell=1}^{N}\mathbf{I}(\mathbf{y^{(\ell)}}=i)}
%\end{equation}

\begin{equation}
\begin{aligned}
&  \mathbf{a}_{ji} = \frac{\sum_{\ell=1}^{N}\sum_{\mathbf{y}^{(\ell)}=i}\mathbf{I}(\hat{\mathbf{y}}_{j}^{(\ell)}=\mathbf{y}^{(\ell)})}
  {\sum_{\ell=1}^{N}\mathbf{I}(\mathbf{y^{(\ell)}}=i)}\\
& \quad\mathbf{u}_{ji} = \frac{\mathbf{a}_{ji} + 10^{-5}}{\mathbf{a}_{1i}+\mathbf{a}_{2i}+\mathbf{a}_{3i}+10^{-5}}
\end{aligned}
\end{equation}
where $\mathbf{a}_{ji}$ is the $i$-th class accuracy of the $j$-th output, and $\mathbf{I}$ is an indicator function, whose value equals to $1$ when the condition exists and $0$ otherwise. Secondly, for the $t$-th test sample, we are able to obtain three output values $\hat{\mathbf{y}}_{1}^{(t)}$, $\hat{\mathbf{y}}_{2}^{(t)}$, and $\hat{\mathbf{y}}_{3}^{(t)}$ via a feed-forward propagation. Finally, the output value can be derived by using Equation$~$(\ref{output}).

\begin{table}
  \centering
  \caption{Numbers of training and test samples in each class for the Houston data.}\label{HSNumber}
  \scalebox{0.9}{
  \begin{tabular}{cccc}
  \hline
     Class No. & Class Name & Training & Test\\
     \hline
     \hline
     1 & Healthy grass & 198 & 1053 \\
     2 & Stressed grass & 190 & 1064 \\
     3 & Synthetic grass & 192 & 505 \\
     4 & Tree & 188 & 1056 \\
     5 & Soil & 186 & 1056 \\
     6 & Water & 182 & 143 \\
     7 & Residential & 196 & 1072 \\
     8 & Commercial & 191 & 1053 \\
     9 & Road & 193 & 1059 \\
     10 & Highway & 191 & 1036 \\
     11 & Railway & 181 & 1054 \\
     12 & Parking lot 1 & 192 & 1041 \\
     13 & Parking lot 2 & 184 & 285 \\
     14 & Tennis court & 181 & 247 \\
     15 & Running track & 187 & 473 \\
     \hline
     \hline
      -  & Total & 2832 & 12197 \\
     \hline
   \end{tabular}
   }
\end{table}

\section{Experiments}
\subsection{Data Description}
\begin{table}
  \centering
  \caption{Numbers of training and test samples in each class for the Trento data.}\label{TrNumber}
  \scalebox{0.9}{
  \begin{tabular}{cccc}
  \hline
     Class No. & Class Name & Training & Test\\
     \hline
     \hline
     1 & Apple trees & 129 & 3905 \\
     2 & Buildings & 125 & 2778 \\
     3 & Ground & 105 & 374 \\
     4 & Wood & 154 & 8969 \\
     5 & Vineyard & 184 & 10317 \\
     6 & Roads & 122 & 3252 \\
     \hline
     \hline
      -  & Total & 819 & 29595 \\
     \hline
   \end{tabular}
   }
\end{table}

\begin{figure}
  \centering
  \includegraphics[scale = 0.45]{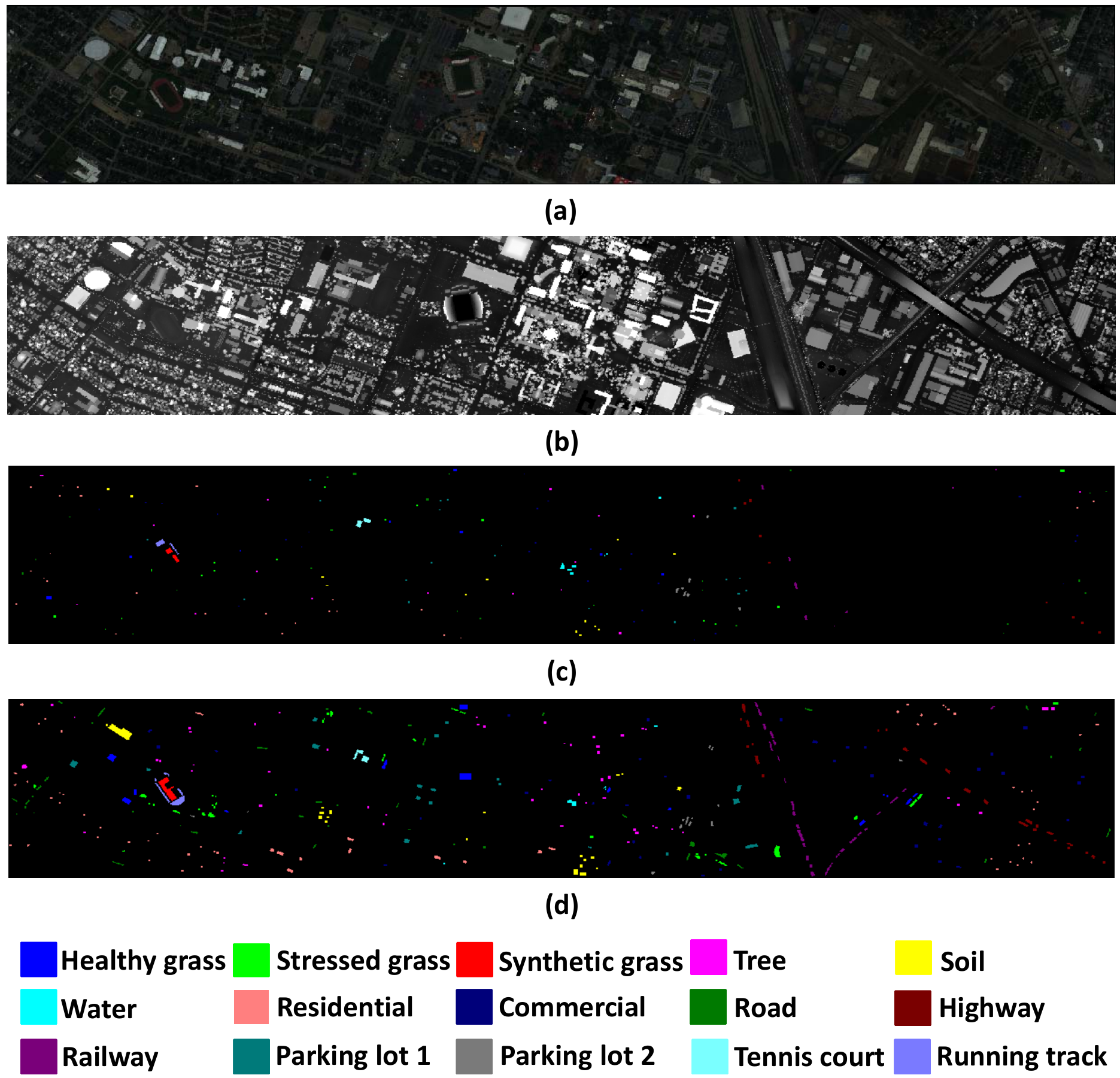}\\
  \caption{Visualization of the Houston data: (a) a pseudo-color image for the hyperspectral data using 64, 43, and 22 as R, G, B, respectively, (b) a grayscale image for the LiDAR data, (c) the training data map, and (d) the test data map.}\label{HSData}
\end{figure}

\begin{figure}
  \centering
  \includegraphics[scale = 0.52]{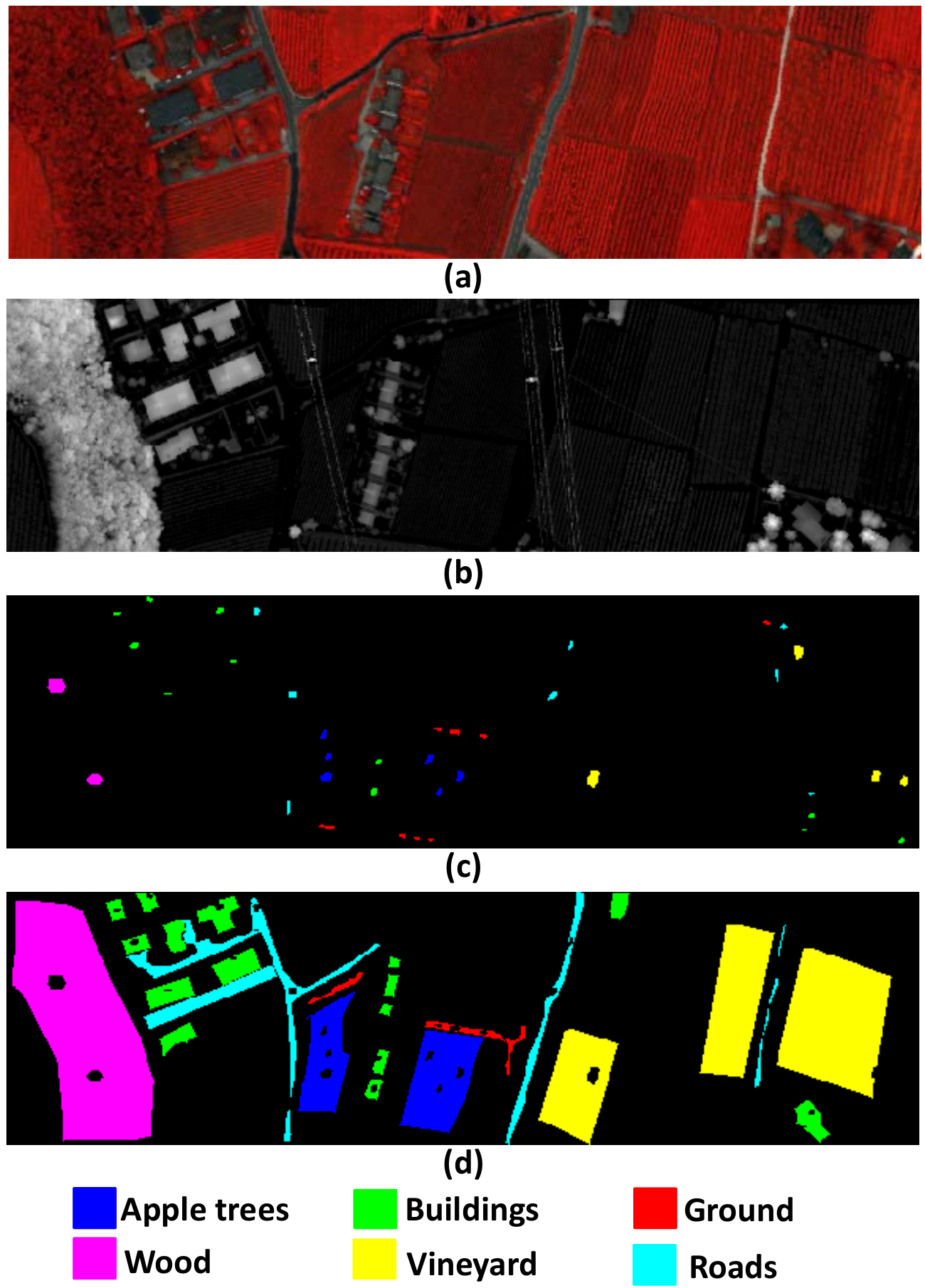}\\
  \caption{Visualization of the Trento data: (a) a pseudo-color image for the hyperspectral data using 40, 20, and 10 as R, G, B, respectively, (b) a grayscale image for the LiDAR data, (c) the training data map, and (d) the test data map.}\label{TrData}
\end{figure}

We test the effectiveness of our proposed model on two hyperspectral and LiDAR fusion data sets.

1) \textbf{Houston Data}: The first data was acquired over the University of Houston campus and the neighboring urban area on June, 2012 \cite{debes2014}. It consists of a hyperspectral image and LiDAR data, both of which contain $349\times 1905$ pixels with a spatial resolution of 2.5 m. The number of spectral bands for the hyperspectral data is 144. Fig.$~$\ref{HSData} demonstrates a pseudo-color image of the hyperspectral data, a grayscale image of the LiDAR data, and ground-truth maps of the training and test samples. As shown in the figure, there exist 15 different classes. The detailed numbers of samples for each class are reported in Table$~$\ref{HSNumber}. It is worth noting that we use the standard sets of training and test samples which makes our results fully comparable with several works such as \cite{ghamisi2018multisource} and \cite{debes2014}.

2) \textbf{Trento Data}: The second data was captured over a rural area in the south of Trento, Italy. The LiDAR data was acquired by the Optech ALTM 3100EA sensor, and the hyperspectral data was acquired by the AISA Eagle sensor with 63 spectral bands. The size of these two data is $166\times 600$ pixels, and the spatial resolution is 1 m. Fig.$~$\ref{TrData} visualizes this data, and Table$~$\ref{TrNumber} lists the number of samples in 6 different classes. Again, we also use the standard sets of training and test samples to construct experiments.

\subsection{Experimental Setup}
In order to validate the effectiveness of our proposed models, we comprehensively compare it with several different models. Specifically, we first select the HS network (i.e., CNN-HS) and the LiDAR network (i.e., CNN-LiDAR) in Fig.$~$\ref{Framework} as two baselines, and compare different fusion methods on both Houston and Trento data. Then, we focus on the Houston data, and compare our model with numerous state-of-the-art models.

%In order to validate the effectiveness of our proposed models, we select the HS network and the LiDAR network in Fig.$~$\ref{Framework} as two baselines. For simplicity, they are abbreviated as CNN-HS and CNN-LiDAR, respectively. Besides, we also test the effectiveness of feature-level fusion models, i.e., using $f_{3}$ only. The three feature-level fusion methods CNN-F-C, CNN-F-M, and CNN-F-S stand for the concatenation method, the maximization method, and the summation method, respectively. Similarly, the three decision-level and feature-level fusion methods in Fig.$~$\ref{Fusion} are abbreviated as CNN-DF-C, CNN-DF-M, and CNN-DF-S, respectively.

All of the deep learning models are implemented in the PyTorch framework. To optimize them, we use the Adam algorithm. The batch size, the learning rate, and the number of training epochs are set to 64, 0.001, and 200, respectively. The experiments are implemented
on a personal computer with an Intel core i7-4790, 3.60GHz processor, 32GB RAM, and a GTX TITAN X graphic card.

The classification performance of each model is evaluated by the overall accuracy (OA), the average accuracy (AA), the per-class accuracy, and the Kappa coefficient. OA defines the ratio between the number of correctly classified pixels to the total number of pixels in the test set, AA refers to the average of accuracies in all classes, and Kappa is the percentage of
agreement corrected by the number of agreements that would be expected purely by chance.

\subsection{Experimental Results}
\begin{table*}
  \centering
  \caption{Classification accuracies (\%) and Kappa coefficients of different models on the Houston data. The best accuracies are shown with the bold type face.}\label{HSResults}
  \scalebox{0.9}{
  \begin{tabular}{ccccccccc}
    \hline
    Class No. & CNN-HS & CNN-LiDAR & CNN-F-C & CNN-F-M & CNN-F-S & CNN-DF-C & CNN-DF-M & CNN-DF-S \\
    \hline
    \hline
    1 & 82.91 & 60.30 & 82.91 & 81.86 & 89.93 & 82.81 & 83.00 & 85.57 \\
    2 & 99.91 & 24.34 & 99.81 & 99.44 & 98.21 & 100 & 99.81 & 99.81	\\
    3 & 91.29 & 66.53 & 97.43 & 97.03 & 98.61 & 96.44 & 97.62 & 97.62\\
    4 & 95.93 & 88.73 & 99.43 & 99.05 & 99.05 & 98.96 & 99.91 & 99.43 \\
    5 & 100 & 24.81 & 100 & 98.86 & 99.72 & 100 & 99.91 & 100 \\
    6 & 93.71 & 25.87 & 96.50 & 100 & 100 & 100 & 100 & 95.80 \\
    7 & 91.60 & 61.19 & 87.41 & 96.74 & 91.98 & 91.32 & 90.39 & 95.24 \\
    8 & 87.18 & 84.33 & 91.17 & 92.69 & 96.30 & 92.40 & 95.54 & 96.39 \\
    9 & 86.87 & 40.32 & 87.25 & 92.92 & 92.92 & 89.33 & 93.86 & 93.20 \\
    10 & 97.59 & 53.86 & 98.75 & 84.94 & 88.51 & 99.71 & 96.04 & 98.84 \\
    11 & 89.56 & 80.46 & 97.15 & 97.34 & 96.49 & 99.43 & 98.39 & 96.77 \\
    12 & 91.16 & 29.30 & 96.25 & 92.22 & 86.65 & 92.51 & 93.18 & 92.60 \\
    13 & 88.77 & 81.05 & 92.98 & 92.63 & 89.82 & 89.82 & 92.98 & 92.98 \\
    14 & 89.07 & 52.63 & 93.52 & 100 & 99.60 & 88.26 & 95.95 & 99.19 \\
    15 & 90.91 & 29.81 & 100 & 92.81 & 99.58 & 100 & 98.73 & 100 \\
    \hline
    \hline
    OA & 92.05 & 54.52 & 94.37 & 93.92 & 94.49 & 94.74 & 95.29 & \textbf{96.03} \\
    AA & 91.76 & 53.57 & 94.70 & 94.57 & 95.16 & 94.73 & 95.69 & \textbf{96.23} \\
    Kappa & 0.9136 & 0.5082 & 0.9389 & 0.9340 & 0.9402 & 0.9429 & 0.9488 & \textbf{0.9569} \\
    \hline
  \end{tabular}
  }
\end{table*}

\begin{figure*}
  \centering
  \includegraphics[scale=0.6]{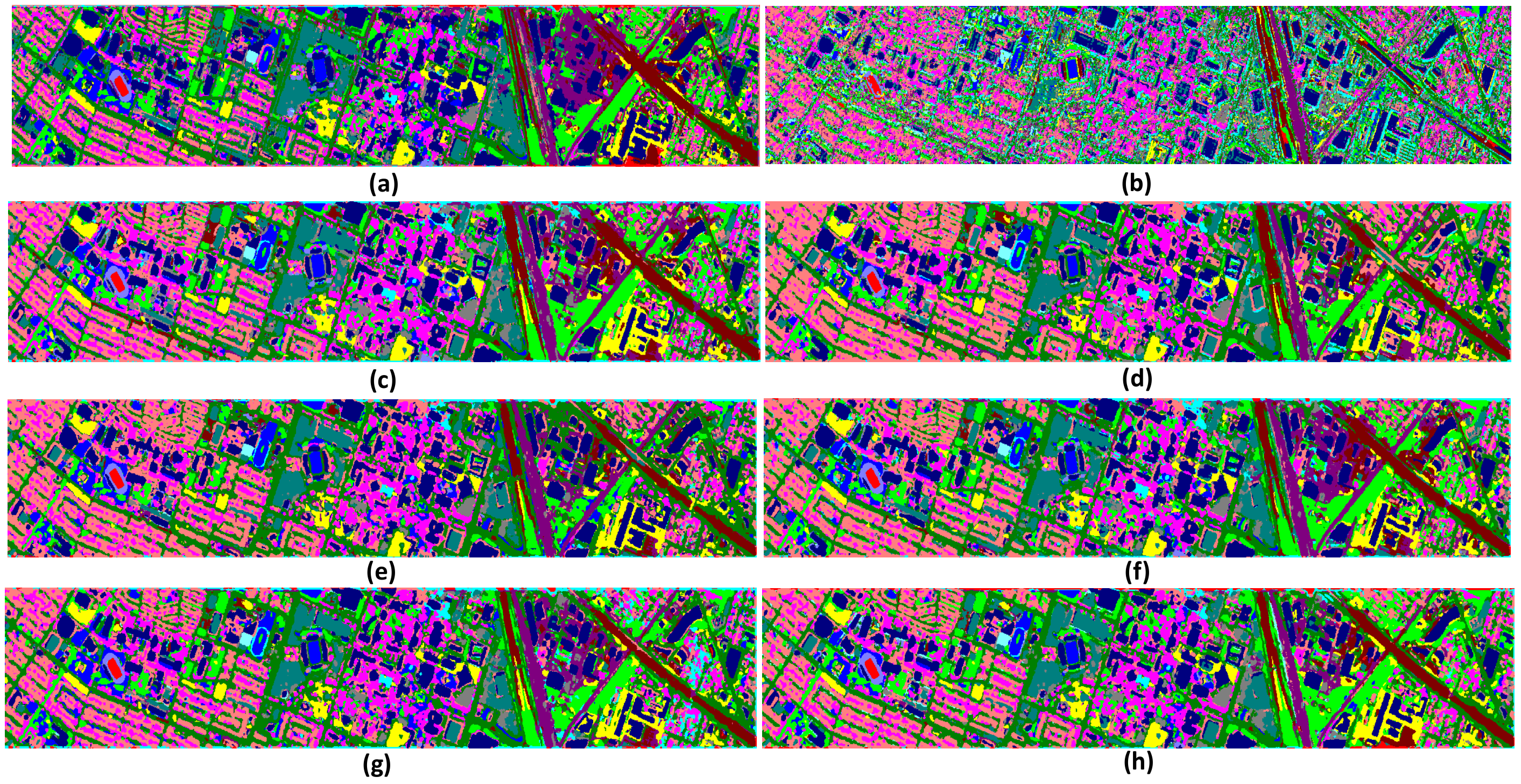}\\
  \caption{Classification maps of the Houston data using different models: (a) CNN-HS, (b) CNN-LiDAR, (c) CNN-F-C, (d) CNN-F-M, (e) CNN-F-S, (f) CNN-DF-C, (g) CNN-DF-M, (h) CNN-DF-S. }\label{HSClaMap}
\end{figure*}

\begin{table*}
  \centering
  \caption{Classification accuracies (\%) and Kappa coefficients of different models on the Trento data. The best accuracies are shown with the bold type face.}\label{TrResults}
  \scalebox{0.9}{
  \begin{tabular}{ccccccccc}
    \hline
    Class No. & CNN-HS & CNN-LiDAR & CNN-F-C & CNN-F-M & CNN-F-S & CNN-DF-C & CNN-DF-M & CNN-DF-S \\
    \hline
    \hline
    1 & 99.85 & 99.92 & 98.49 & 96.72 & 99.15 & 98.44 & 99.69 & 99.64 \\
    2 & 94.67 & 93.16 & 97.01 & 97.05 & 96.36 & 97.73 & 98.81 & 97.66 \\
    3 & 82.09 & 60.43 & 92.51 & 95.99 & 93.05 & 88.50 & 94.39 & 92.25 \\
    4 & 98.73 & 99.12 & 99.11 & 100 & 100 & 100 & 99.88 & 99.96 \\
    5 & 99.73 & 95.63 & 100 & 100 & 99.96 & 100 & 100 & 99.90 \\
    6 & 76.31 & 50.59 & 90.53 & 92.69 & 89.71 & 93.64 & 94.00 & 92.40 \\
    \hline
    \hline
    OA & 96.31 & 91.91 & 98.17 & 98.48 & 98.37 & 98.77 & \textbf{99.12} & 98.80 \\
    AA & 91.90 & 83.14 & 96.28 & 97.08 & 96.37 & 96.39 & \textbf{97.80} & 96.97 \\
    Kappa & 0.9505 & 0.8917 & 0.9754 & 0.9796 & 0.9782 & 0.9835 & \textbf{0.9881} & 0.9839 \\
    \hline
  \end{tabular}
  }
\end{table*}

\begin{figure*}
  \centering
  \includegraphics[scale=0.6]{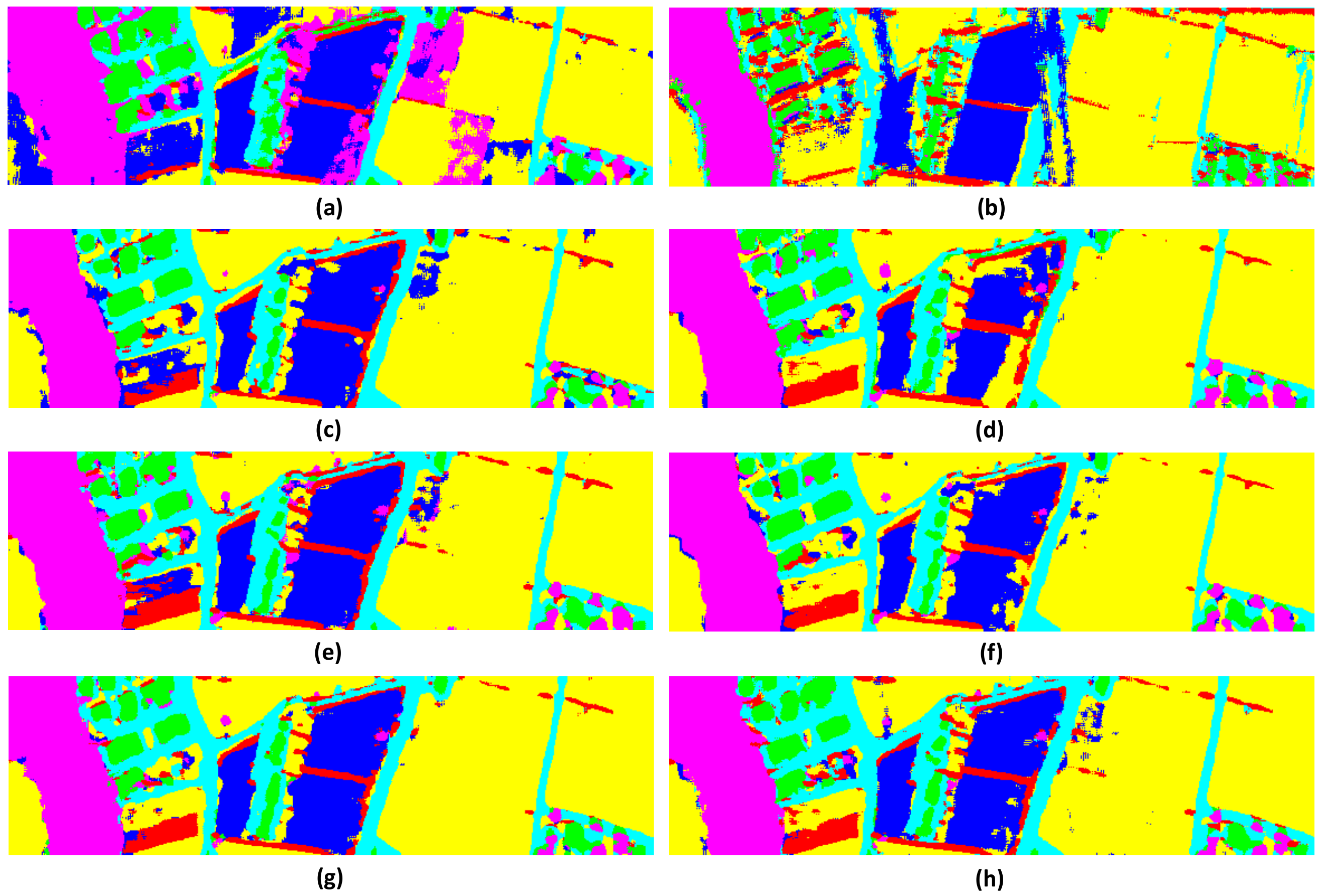}\\
  \caption{Classification maps of the Trento data using different models: (a) CNN-HS, (b) CNN-LiDAR, (c) CNN-F-C, (d) CNN-F-M, (e) CNN-F-S, (f) CNN-DF-C, (g) CNN-DF-M, (h) CNN-DF-S. }\label{TrClaMap}
\end{figure*}

\begin{table*}
\centering
\caption{Performance comparison with the state-of-the-art models on the Houston data. }\label{Compare}
\scalebox{0.9}{
\begin{tabular}{cccccccc}
\hline
\multicolumn{7}{c}{\textbf{Traditional models}}                          \\
\hline
Model & MLR$_{sub}$ & GGF & SLRCA & OTVCA & ODF-ADE & E-UGF & HyMCKs \\
OA  & 92.05 & 94.00 & 91.30 & 92.45 & 93.50 & 95.11 & 90.33  \\
AA & 92.87 & 93.79 & 91.95 & 92.68 & - & 94.57 & 91.14 \\
Kappa & 0.9137 & 0.9350 & 0.9056 & 0.9181 & 0.9299 & 0.9447 & 0.8949\\
\hline
\hline
\multicolumn{7}{c}{\textbf{CNN-related models}}                           \\
\hline
Model & DF & CNNGBFF & CNNCK & TCNN & PToPCNN & CNN-DF-M & CNN-DF-S   \\
OA    & 91.32 & 91.02 & 92.57 & 87.98 & 92.48 & \textbf{95.29} & \textbf{96.03} \\
AA    & 91.96 & 91.82 & 92.48 & 90.11 & 93.55 & \textbf{95.69} & \textbf{96.23} \\
Kappa & 0.9057 & 0.9033 & 0.9193 & 0.8698 & 0.9187 & \textbf{0.9488} & \textbf{0.9569} \\
\hline
\end{tabular}
}
\end{table*}

\subsubsection{Comparison with different fusion models}
In addition to two single-source models (i.e., CNN-HS and CNN-LiDAR), we also test the effectiveness of feature-level fusion models, i.e., using $f_{3}$ only. The three feature-level fusion methods CNN-F-C, CNN-F-M, and CNN-F-S stand for the concatenation method, the maximization method, and the summation method, respectively. Similarly, the three decision-level and feature-level fusion methods in Fig.$~$\ref{Fusion} are abbreviated as CNN-DF-C, CNN-DF-M, and CNN-DF-S, respectively.
Table$~$\ref{HSResults} shows the detailed classification results of eight models on the Houston data. Several conclusions can be observed from it. First, for the single-source models, CNN-HS achieves significantly better results than CNN-LiDAR in each class. It indicates that the spectral-spatial information in the hyperspectral data is more discriminative than the elevation information in the LiDAR data. Second, all of the three feature-level fusion models (i.e., CNN-F-C, CNN-F-M, and CNN-F-S) obtain higher accuracies than the CNN-HS model in most classes. This can be explained by that the LiDAR data can provide complementary information for the hyperspectral data, and by combining them together in a proper way, the classification performance can be improved. Third, based on the feature-level fusion models, if we further use the decision-level fusion (i.e., CNN-DF-C, CNN-DF-M, and CNN-DF-S), the performance is improved again. Taking the summation fusion method as an example, by the simultaneous use of feature-level and decision-level fusions, the OA is increased from 94.49\% to 96.03\%, which is the best result ever reported in the literature. Last but not the least, compared to the widely used concatenation method, our proposed maximization and summation fusion methods can achieve better OA, AA, and Kappa values. Besides the quantitative results, we also qualitatively analyze the performance of different models. Fig.$~$\ref{HSClaMap} demonstrates the classification maps of different models. In this figure, different colors represent different classes of objects. From Fig.$~$\ref{HSClaMap}(b), we can see that the CNN-LiDAR model generates many outliers, and misclassifies a lot of objects. In comparison with it, other models obtain more homogeneous classification maps. However, some objects are a little over-smoothed, because all of the models use the small patches and cubes as inputs.

Similar to the Houston data, Table$~$\ref{TrResults} and Fig.$~$\ref{TrClaMap} show the quantitative and qualitative results, respectively, on the Trento data. The data have larger and more homogeneous objects to discriminate than the Houston data, so all of the models can achieve relatively high performance (e.g., the OA values are larger than 90\%). Specifically, CNN-HS is better than CNN-LiDAR, and the feature-level fusion method can improve the performance of CNN-HS. More importantly, the simultaneous feature-level and decision-level fusion is more effective than using feature-level fusion only. The best results appear when adopting the maximization fusion method.

%\begin{table*}
%\centering
%\caption{Performance comparison with the state-of-the-art models on the Trento data.}\label{PatchSize}
%\scalebox{0.9}{
%\begin{tabular}{cccccccc}
%\hline
%\multicolumn{7}{c}{\textbf{Traditional models}}                          \\
%\hline
%Model & MLR$_{sub}$ & GGF & SLRCA & OTVCA & ODF-ADE & E-UGF & HyMCKs \\
%OA  & - & - & 99.27 & 99.48 & - & - & 98.97  \\
%AA & - & - & 98.55 & 99.08 & - & - & 98.18\\
%Kappa & - & - & 0.9902 & 0.9931 & - & - & 0.9863\\
%\hline
%\hline
%\multicolumn{7}{c}{\textbf{CNN-related models}}                           \\
%\hline
%Model & DF & CNNGBFF & CNNCK & TCNN & PToPCNN & CNN-DF-M & CNN-DF-S   \\
%OA    & - & 98.93 & 97.33 & 97.92 & 98.34 & 99.12 & 98.80 \\
%AA    & - & 98.48 & 92.19 & 96.19 & 97.53 & 97.80 & 96.97\\
%Kappa & - & 0.9855 & 0.9063 & 0.9681 & 0.9779 & 0.9881 & 0.9839\\
%\hline
%\end{tabular}
%}
%\end{table*}

\subsubsection{Comparison with state-of-the-art models}
In the existing hyperspectral and LiDAR data fusion works, most of models tested their performance on the Houston data. To highlight the superiority of our proposed models, we also compare them with state-of-the-art models, including 7 traditional models and 5 CNN-related models, using standard training and test sets. These traditional models include the multiple feature learning model MLR$_{sub}$ in \cite{khodadadzadeh2015}, the generalized graph-based fusion model GGF in \cite{liao2015}, the sparse and low-rank component analysis model SLRCA in \cite{rasti2017fusion}, the total variation component analysis model OTVCA in \cite{rasti2017}, the adaptive differential evolution based fusion model ODF-ADE in \cite{zhong2017}, the unsupervised graph fusion model E-UGF in \cite{xia2018}, and the composite kernel extreme learning machine model HyMCKs in \cite{ghamisi2019}. The CNN-related models include the deep fusion model DF in \cite{chen2017deep}, the CNN model combined with graph-based feature fusion method CNNGBFF in \cite{ghamisi20172}, the three-stream CNN based  composite kernel model CNNCK in \cite{li2018}, the two-branch CNN model TCNN in \cite{xu2018}, and the patch-to-patch CNN model PToPCNN in \cite{zhang2018}.

Table$~$\ref{Compare} reports the detailed comparison results of different models in terms of OA, AA, and Kappa coefficients. Note that all the results are directly cited from their original papers, because we are not able to reproduce them due to missing parameters
or availability of codes. For the traditional models, the best OA, AA, and Kappa values are 95.11\%, 94.57\%, and 0.9447, respectively, achieved by a recent work named E-UGF \cite{xia2018}. For the CNN-related models, CNNCK \cite{li2018} obtains the best OA and Kappa values, while PToPCNN \cite{zhang2018} acquires the best AA. Compared to the E-UGF model, both CNNCK and PToPCNN models obtain inferior performance, which indicate that the existing CNN-related fusion models still have some potentials to explore. Similar to DF \cite{chen2017deep} and TCNN \cite{xu2018} models, our proposed models (i.e., CNN-DF-M and CNN-DF-S) can also be considered as a two-branch CNN model. However, the proposed models can obtain significantly better results than them, even than E-UGF, which sufficiently certify the effectiveness of the proposed model.

\subsection{Analysis on the proposed model}
\subsubsection{Analysis on the reduced dimensionality}
\begin{figure}
  \centering
  \includegraphics[scale=0.5]{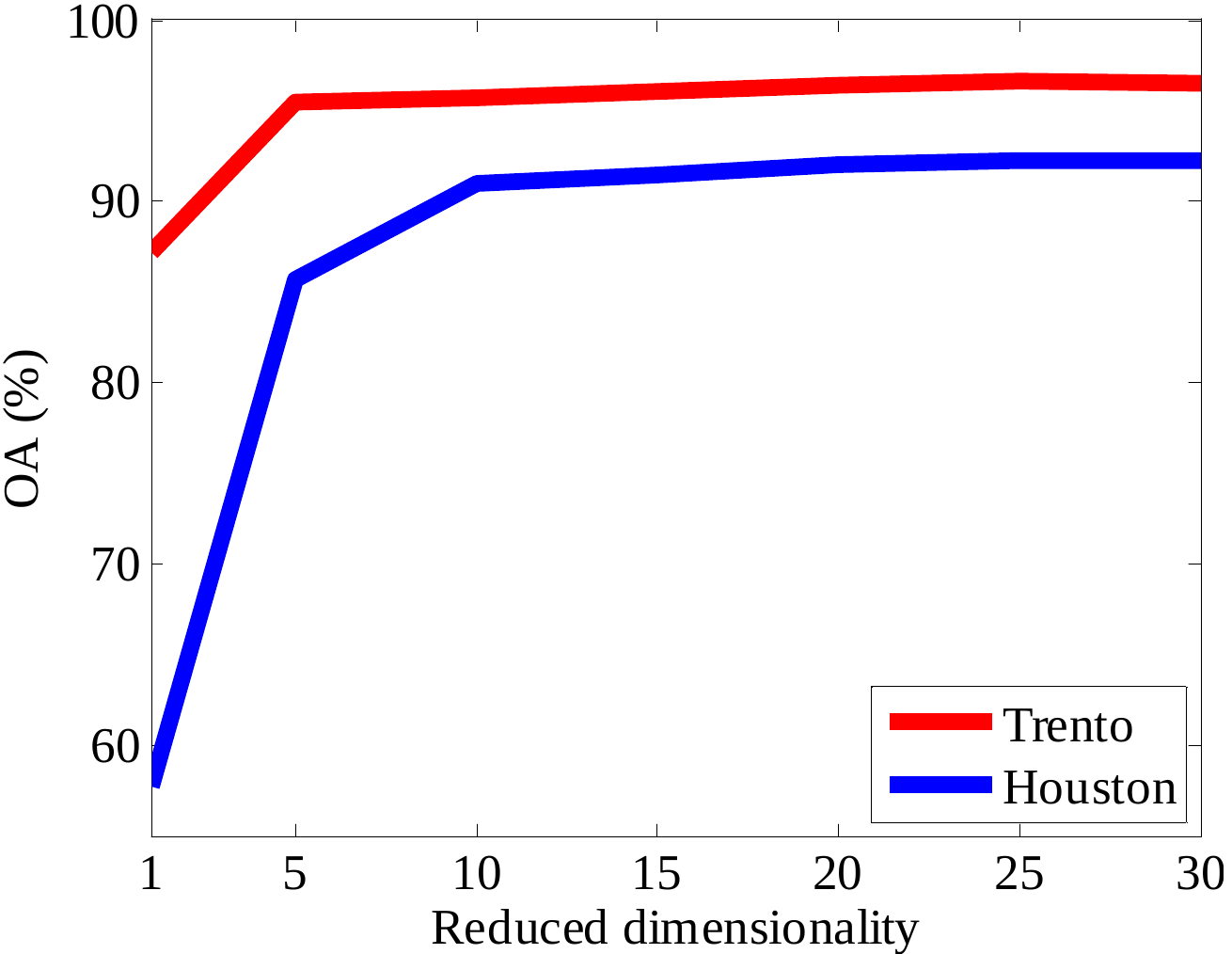}\\
  \caption{Effect of the reduced dimensionality on the OA (\%) achieved by the CNN-HS model.}\label{Dim}
\end{figure}

For the proposed model, we have two hyper-parameters to predefine. The first one is the number of reduced dimensionality $k$ of hyperspectral data using PCA, and the second one is the neighboring size $p\times p$ extracted from hyperspectral and LiDAR data. To evaluate the effect of $k$, we fix $p$ and select $k$ from a candidate set $\{1,5,10,15,20,25,30\}$. Since the fusion models have the same hyper-parameter values as single models (i.e., CNN-HS and LiDAR-HS), we only demonstrate the results of single models here. Fig.$~$\ref{Dim} shows the performance (i.e., OA) of CNN-HS on the Houston (the blue line) and Trento (the red line) data. From this figure, we can observe that as $k$ increases, OA firstly increases and then tends to a stable state. Considering the computation complexity and classification performance, $k$ can be set to 20 for both data.

\subsubsection{Analysis on the neighboring size}
Similar to the analysis of $k$, we can also fix $k$ and choose $p$ from a candidate set $\{9,11,13,15,17,19\}$ to evaluate the effect of $p$. Table$~$\ref{PatchSize} reports the changes of OA values at different sizes. When the size increases from 9 to 11 on the Houston data, the improvements of OA acquired by CNN-HS and CNN-LiDAR are more than 1 percent. But for the other sizes, these two models do not change significantly. For the Trento data, CNN-HS is relatively stable when the size changes, but CNN-LiDAR will increase more than 1 percent from 9 to 11, and decrease from 11 to 13. Based on the above analysis, 11 is a reasonable choice for CNN-HS and CNN-LiDAR on both data. This choice is consistent with the works in \cite{chen2017deep} and \cite{zhang2018}.

\begin{table}
\centering
\caption{Effect of the neighboring size on the OA (\%) acquired by the CNN-HS and CNN-LiDAR models.}\label{PatchSize}
\scalebox{0.9}{
\begin{tabular}{ccccccc}
\hline
\multicolumn{7}{c}{\textbf{Houston Data}}                          \\
\hline
Size & 9 & 11 & 13 & 15 & 17 & 19 \\
CNN-HS    & 90.88 & 92.05 & 91.49 & 91.41 & 91.87 & 92.06 \\
CNN-LiDAR & 52.45 & 54.52 & 54.44 & 54.59 & 54.29 & 54.51 \\
\hline
\hline
\multicolumn{7}{c}{\textbf{Trento Data}}                           \\
\hline
Size      & 9     & 11    & 13    & 15    & 17    & 19    \\
CNN-HS    & 96.02 & 96.43 & 96.39 & 96.17 & 95.97 & 95.53 \\
CNN-LiDAR & 90.80 & 91.91 & 90.29 & 90.70 & 91.40 & 90.57 \\
\hline
\end{tabular}
}
\end{table}

\begin{table}
\centering
\caption{Computation time (seconds) of different models on the Houston data.}\label{HSTime}
\scalebox{0.9}{
\begin{tabular}{ccccc}
%\hline
%\multicolumn{7}{c}{\textbf{Houston Data}}                          \\
\hline
Time & CNN-HS & CNN-LiDAR & CNN-F-C & CNN-F-M \\
Train  & 43.68 & 38.04  & 71.57 & 70.85  \\
Test & 1.24 & 1.18 & 1.30 & 1.27 \\
\hline
\hline
%\multicolumn{7}{c}{\textbf{Trento Data}}                           \\
%\hline
Time      & CNN-F-S & CNN-DF-C & CNN-DF-M & CNN-DF-S   \\
Train    & 70.90 & 185.71 & 182.54 & 184.43\\
Test  & 1.28 & 1.38& 1.33 & 1.37 \\
\hline
\end{tabular}
}
\end{table}

\begin{table}
\centering
\caption{Computation time (seconds) of different models on the Trento data.}\label{TrTime}
\scalebox{0.9}{
\begin{tabular}{ccccc}
%\hline
%\multicolumn{7}{c}{\textbf{Houston Data}}                          \\
\hline
Time & CNN-HS & CNN-LiDAR & CNN-F-C & CNN-F-M \\
Train  & 32.11 & 21.84 & 49.99 & 49.53 \\
Test & 1.33 & 1.24 & 1.44 & 1.37\\
\hline
\hline
%\multicolumn{7}{c}{\textbf{Trento Data}}                           \\
%\hline
Time      & CNN-F-S & CNN-DF-C & CNN-DF-M & CNN-DF-S   \\
Train    & 49.62 & 118.65 & 116.43 & 117.29 \\
Test  & 1.43 & 1.66 & 1.62 & 1.65 \\
\hline
\end{tabular}
}
\end{table}

\begin{figure}
  \centering
  \includegraphics[scale = 0.3]{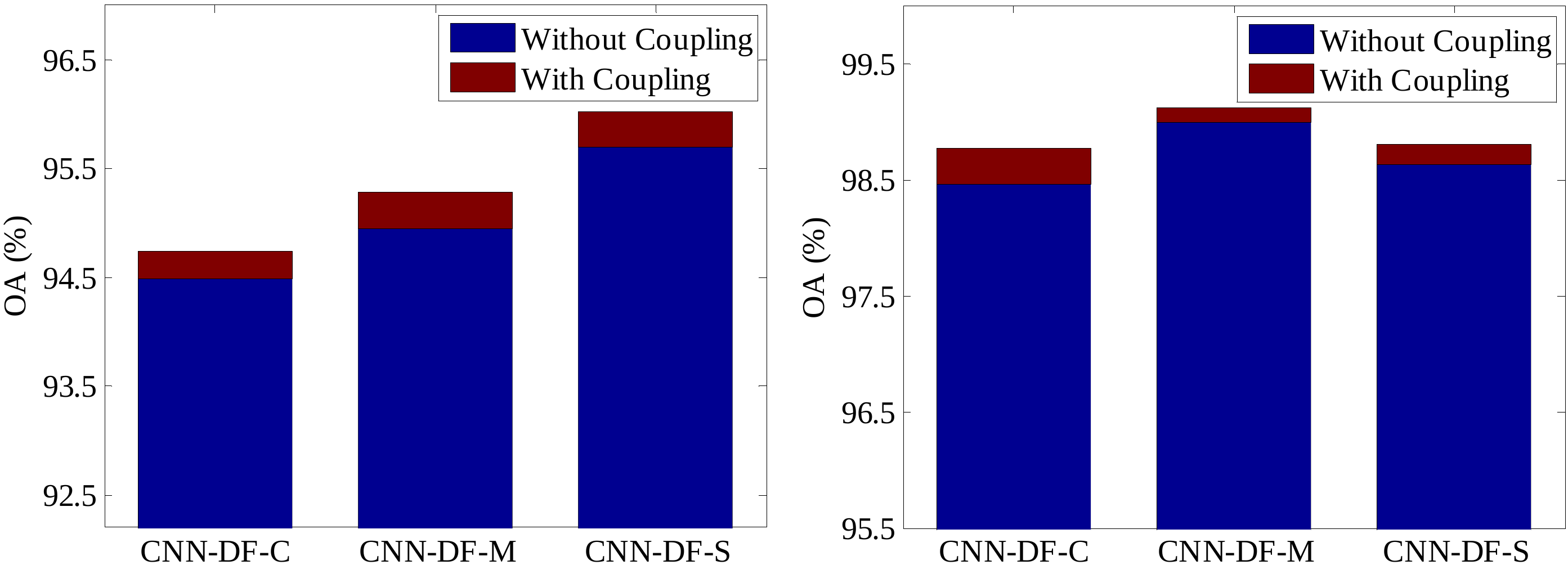}\\
  \caption{Comparisons before and after adopting the coupling strategy on two data. From left to right is the Houston data and the Trento data.}\label{EffectsofCoupling}
\end{figure}
\subsubsection{Analysis on the coupling strategy}
Benefiting from the coupling strategy, the number of parameters in the second and the third convolutional layers is reduced by twice. Taking CNN-DF-M and CNN-DF-S models as an example, on the Houston data, the total number of parameters to train is 196128 without weight sharing, while this number is reduced to 103968 after adopting the coupling strategy; on the Trento data, the trainable parameters are 192672 and 100512 without and with weight sharing, respectively. In summary, the parameter numbers in CNN-DF-M and CNN-DF-S models are reduced by about 47\% on both data when the coupling strategy is employed. Besides, we also test the effects of the coupling strategy on the classification performance. Fig.$~$\ref{EffectsofCoupling} illustrates the changes of OA before and after adopting the coupling strategy on the Houston data (left one) and the Trento data (right one). It indicates that the performance of CNN-DF-C, CNN-DF-M, and CNN-DF-S in terms of OA is slightly improved after adopting the coupling strategy.

\subsubsection{Analysis on the computation cost}
To quantitatively analyze the computation cost of different models, Table$~$\ref{HSTime} and Table$~$\ref{TrTime} report their computation time on the Houston and Trento data, respectively. From these two tables, we can observe that CNN-HS and CNN-LiDAR models take less training time than the other fusion models, because they only need to process single-source data, without any interactions between different sources. On the contrary, the proposed decision-level and feature-level fusion models cost much more training time than the single-source and the feature-level fusion models. Nevertheless, once the networks are trained, their test efficiency is very high. In particular, it takes no more than 2 seconds to finish the test process, which is close to the time costs of the other models.

\begin{figure}
  \centering
  \includegraphics[scale = 0.35]{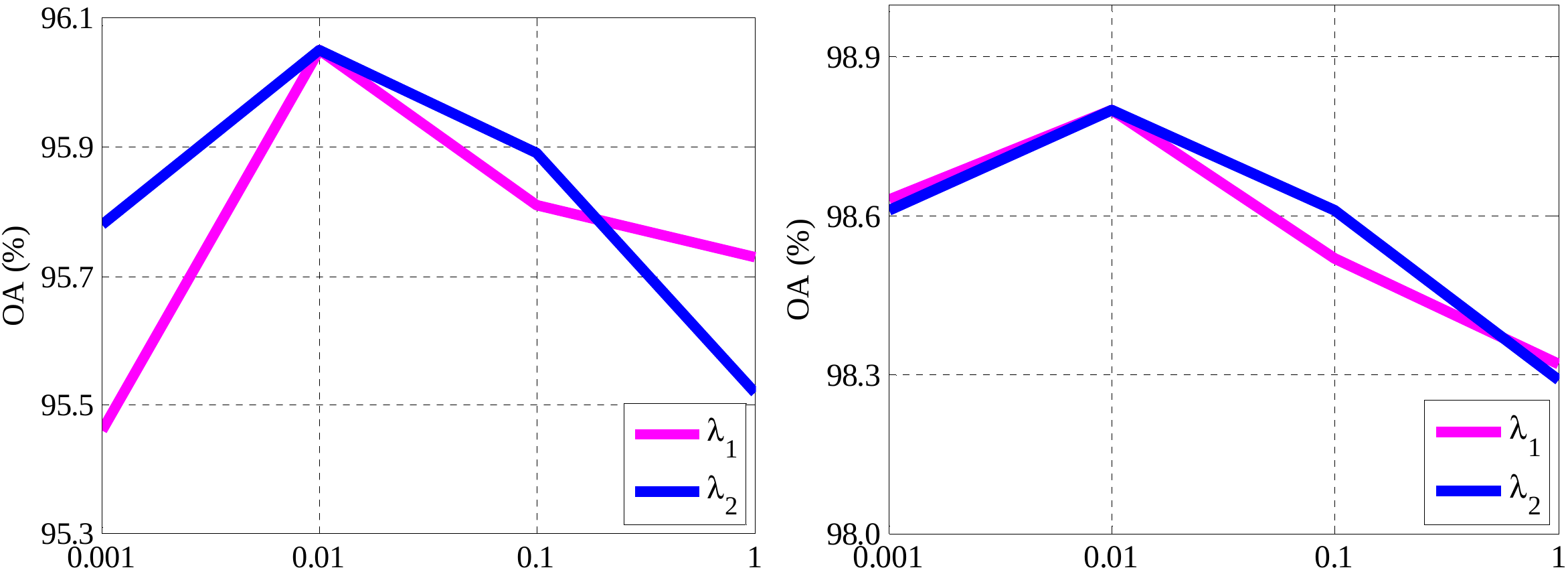}\\
  \caption{Effects of weight parameters $\lambda_{1}$ and $\lambda_{2}$ on the classification performance achieved by the CNN-DF-S model on two data. From left to right is the Houston data and the Trento data.}\label{EffectsofLambda}
\end{figure}
\subsubsection{Analysis on the weight parameters}
The loss function of the proposed model in Equation$~$(\ref{Loss}) contains two hyper-parameters (i.e., $\lambda_{1}$ and $\lambda_{2}$). In order to test their effects on the classification performance, we firstly fix $\lambda_{1}$ and change $\lambda_{2}$ from a candidate set $\{0.001, 0.01, 0.1, 1\}$. Then, we set $\lambda_{2}$ to the optimal value and change $\lambda_{1}$ from the same set $\{0.001, 0.01, 0.1, 1\}$. Fig.$~$\ref{EffectsofLambda} shows the OAs obtained by the proposed CNN-DF-S model on the Houston and Trento data with different $\lambda_{1}$ and $\lambda_{2}$ values. In this figure, the pink and the blue lines represent the CNN-DF-S model with different $\lambda_{1}$ and $\lambda_{2}$ values, respectively. It is shown that as $\lambda_{2}$ increases, the OA will firstly increase and then decrease on both data. The highest OA value appears when $\lambda_{2}=0.01$. Similar conclusions can be observed for $\lambda_{1}$. Therefore, the optimal values for $\lambda_{1}$ and $\lambda_{2}$ are 0.01.

\section{Conclusions}
This paper proposed a coupled CNN framework for hyperspectral and LiDAR data fusion. Small convolution kernels and parameter sharing layers were designed to make the model more efficient and effective. In the fusion phase, we used feature-level and decision-level fusion strategies simultaneously. For the feature-level fusion, we proposed summation and maximization methods in addition to the widely used concatenation method. For the decision-level fusion, we proposed a weighted summation method, whose weights depend on the performance of each output layer. To validate the effectiveness of the proposed model, we constructed several experiments on two data sets. The experimental results show that the proposed model can achieve the best performance on the Houston data, and very high performance on the Trento data. Additionally, we also thoroughly evaluated the effects
of different hyper-parameters on the classification performance, including the reduced dimensionality and the neighboring size. In the future, more powerful neighboring extraction methods need to be explored, because the current classification maps still exist over-smoothing problems.

\bibliography{IEEEfull,LiDARandHSI}
\bibliographystyle{IEEEbib}

\end{document}